\documentclass[journal,10pt]{IEEEtran}
\usepackage[T1]{fontenc}
\usepackage{cite}
\usepackage{amsmath,amssymb,amsfonts}
\usepackage{graphicx}
\usepackage{booktabs,array}
\usepackage{placeins}
\usepackage{algorithm,algorithmic}
\usepackage{xcolor}
\usepackage{hyperref}
\hypersetup{hidelinks,pdftitle={OKB-CFM: Observation-Keyed Barycentric Conditional Flow Matching},pdfauthor={Kangneng Zhou, Shuang Song, Yitian Zhao, Dian Song, Yujia Wang, Xinghua Zhang, Xi Li, Feng Cao}}
\usepackage{microtype}
\makeatletter
\def\section{\@startsection{section}{1}{\z@}
{2.0ex plus 0.2ex minus 0.4ex}{0.5ex plus 0.1ex}
{\normalfont\normalsize\centering\scshape}}
\def\subsection{\@startsection{subsection}{2}{\z@}
{1.5ex plus 0.2ex minus 0.4ex}{0.5ex plus 0.1ex}
{\normalfont\normalsize\itshape}}
\makeatother
\usepackage{etoolbox}
\AtBeginEnvironment{equation}{%
\setlength{\abovedisplayskip}{4pt plus 1pt minus 1pt}%
\setlength{\belowdisplayskip}{4pt plus 1pt minus 1pt}%
\setlength{\abovedisplayshortskip}{0pt plus 1pt}%
\setlength{\belowdisplayshortskip}{3pt plus 1pt minus 1pt}}

\begin{document}
\title{OKB-CFM: Observation-Keyed Barycentric Conditional Flow Matching for Unpaired Sparse-to-Dense Cardiac Shape Completion}
\author{Kangneng~Zhou, Shuang~Song, Yitian Zhao, Dian~Song, Yujia~Wang, \\
Xinghua~Zhang, Li Wang, Xi Li, Hao Dai and~Feng~Cao
\thanks{This work was supported in part by grants from Key Research Program of Beijing Natural Science Foundation (L241026) and Fundamental Research Funds for the Central Universities (FRF-TP-24-022A). Access to UK Biobank data was granted through application number 87841. (Corresponding authors: Shuang Song and Xi Li.)}.
\thanks{Kangneng Zhou, Shuang Song and Dian Song are with School of Computer and Communication Engineering, University of Science and Technology Beijing, Beijing 100083, China (e-mail:elliszkn@163.com; ssong@ustb.edu.cn; ssdian1024@gmail.com).}
\thanks{Yitian Zhao is with Ningbo Institute of Materials Technology and Engineering, Chinese Academy of Sciences, Zhejiang 315201, China. (e-mail: yitian.zhao@nimte.ac.cn)}
\thanks{Yujia Wang is with Medical School of Chinese People′s Liberation Army General Hospital, Beijing 100853, China.(e-mail:Nancylinye@163.com)}
\thanks{Feng Cao is with Second Medical Center of Chinese People′s Liberation Army General Hospital, National Clinical Research Center for Geriatric Diseases, Beijing 100853, China. (e-mail:fengcao8828@163.com)}
\thanks{Xinghua Zhang is with Department of Radiology, First Medical Center of the Chinese People′s Liberation Army General Hospital, Beijing 100853, China. (e-mail:doctorzhxh@163.com)}
\thanks{Li Wang, Xi Li and Hao Dai are with Fuwai Hospital, Chinese Academy of Medical Sciences, National Cardiovascular Disease Key Laboratory, Beijing 100037, China. (e-mail: wangl@pumc.edu.cn; xi.li@nccd.org.cn; daihao@fuwai.com)}
}

\maketitle
\begin{abstract}
Completing cardiac segmentations from sparse cine magnetic resonance labels is challenging because acquired planes do not uniquely determine the underlying anatomy and patient-matched dense annotations are rarely available. We propose observation-keyed barycentric conditional flow matching for completion between clinically unpaired cohorts. Independently trained variational autoencoders represent sparse observations and dense shapes. A fixed memory associates simulated sparse keys with their source dense encodings, enabling candidate retrieval within the sparse latent space. A matching network aggregates the retrieved values into a query-dependent barycentric endpoint, toward which an endpoint-conditioned flow transports the sparse encoding. Individual candidate values provide alternative completion endpoints. On 327 real unpaired inputs, the method obtains encoder-feature Fr\'{e}chet and squared maximum mean discrepancy scores of 7.93 and 2.02 after scaling by 10,000 and 10,000,000, respectively, corresponding to reductions of 47.0\% and 94.2\% relative to contrastive unpaired translation. On a 34-file simulated test partition, evaluation against dense-VAE-reconstructed references yields a five-class mean Dice of 0.889, squared Chamfer distance of 6.15 square voxels, 95th-percentile surface distance of 3.88 voxels, and average symmetric surface distance of 1.50 voxels. These results characterize cohort-level feature agreement, observed-label consistency, and reconstruction under the stated reference construction. By complementing routine cine-based functional assessment with inferred
multi-chamber anatomical context, this framework provides a basis for
investigating how chamber morphology and spatial relationships relate
to cardiac function.
\end{abstract}

\section{Introduction}

Routine cine cardiac magnetic resonance imaging (CMR) supports
functional assessment, but its sparse short-axis and long-axis
acquisitions leave multi-chamber anatomy incompletely
observed~\cite{zhang2025towards,ruijsink2020fully}.
Dense whole-heart segmentations from CT or MRI provide broader
anatomical coverage~\cite{kong2022learning}, yet often come from
independent cohorts with different acquisition and annotation
protocols, lacking patient-level correspondence with cine data.
The challenge is therefore to combine the subject-specific
observations from cine CMR with the more complete anatomical
information available in dense CT/MRI segmentations. This requires
learning from these clinically unpaired data sources to compensate
for the limited spatial coverage of cine CMR while preserving
agreement with the observed anatomy.

Existing completion methods commonly obtain training correspondences from paired or simulated partial--complete anatomies. CardiacFlow~\cite{ma2025cardiacflow} uses simulated acquisitions and flow-generated dense shapes, but simulation does not establish the association between a real observation and an independent dense cohort. Cycle-consistent translation~\cite{yang2020unsupervised,bargshady2022application} encourages approximate invertibility despite the ambiguity of completion. Diffusion and flow models~\cite{ozbey2023unsupervised,li2023bbdm,arslan2025self,chadebec2025lbm,yazdani2025flow} offer flexible distributional transport, yet marginal distribution matching alone cannot identify observation-relevant anatomy.

\begin{figure*}[!t]
\centering
\includegraphics[width=0.9\textwidth]{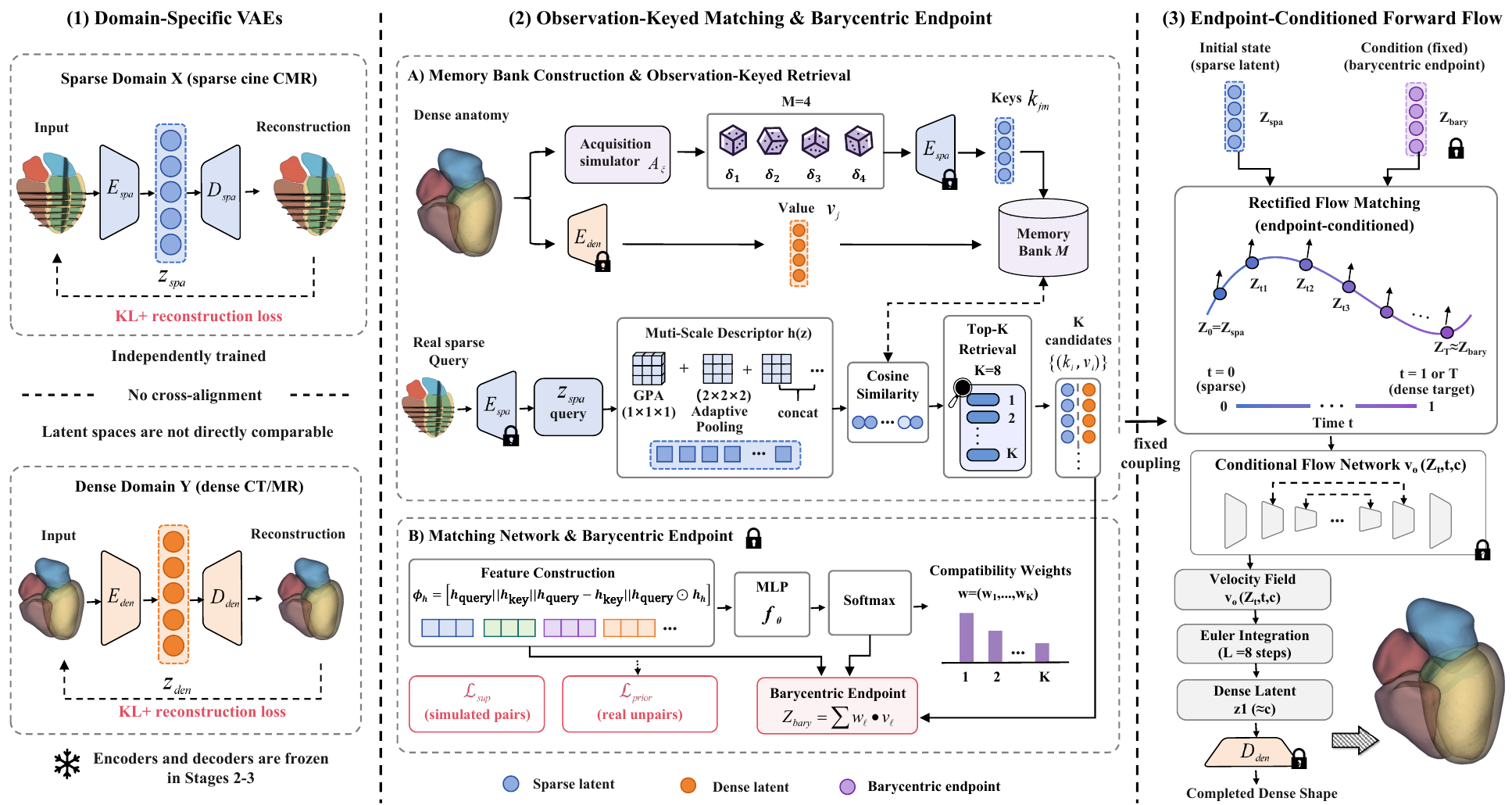}
\caption{Overview of OKB-CFM. Independently trained domain-specific variational autoencoders encode sparse observations and dense segmentations. Sparse-domain retrieval selects observation-keyed dense candidates, and matching aggregates their values into a barycentric endpoint. A conditional velocity field maps the sparse code toward this endpoint for dense decoding. Training is sequential; endpoint conditioning uses global average pooling.}
\label{fig:framework}
\end{figure*}

These limitations motivate constructing an observation-dependent coupling before transport. Independently trained sparse and dense encoders need not share semantic coordinates, making direct cross-domain latent comparison unreliable. Our insight is to link dense shapes to the observation space through simulated sparse acquisitions. OKB-CFM (Fig.~\ref{fig:framework}) implements this association with an observation-keyed memory: retrieval compares a query with sparse keys and returns their source dense encodings. A matcher forms a barycentric endpoint, and an endpoint-conditioned flow transports the sparse code for dense decoding. The clinical cohorts remain unpaired; simulation provides paired observations and source anatomies for training.

Our main contributions are as follows:
\begin{itemize}[\setlength{\IEEElabelindent}{0pt}\setlength{\labelsep}{0.4em}\setlength{\topsep}{0pt}]
\setlength{\itemsep}{0pt}
\setlength{\parsep}{0pt}
\item {Observation-keyed anatomical retrieval.} We introduce a memory linking sparse observation keys to dense shape values, enabling query-dependent candidate selection without assuming semantic alignment between independently trained domain encoders.
\item {Barycentric endpoint construction.} We develop a matching objective combining simulated dense-source supervision, real-query retrieval regularization, and pooled-candidate regularization on simulated barycenters. Learned candidate weights construct an observation-dependent dense endpoint.
\item {Endpoint-conditioned latent transport.} We separate endpoint construction from conditional flow and examine their roles through retrieval, fixed-candidate weighting, direct-decoding, and objective ablations. Individual candidate endpoints additionally support alternative completions.
\end{itemize}

\section{Related Work}

\subsection{Cardiac Shape Completion}

Cardiac reconstruction uses statistical and learned shape
priors~\cite{pandey2025deep}. DeepRecon combines segmentation and
volumetric reconstruction~\cite{chang2022deeprecon}, while mesh-based
and implicit methods represent cardiac geometry through deformable
templates or continuous
functions~\cite{yang2023whole,sander2023reconstruction,muffoletto2026neural}.
These representations constrain shape but do not uniquely determine
anatomy between acquired planes or within incompletely observed chambers.

DeepRecon reconstructs intermediate image slices and their segmentations
by interpolating optimized latent codes of neighboring cine
images~\cite{chang2022deeprecon}.
Neural Implicit Heart Coordinates instead fit a subject-specific latent
code to sparse segmentation labels, using a common ventricular coordinate
system to recover anatomically corresponding
meshes~\cite{muffoletto2026neural}.
Joint image reconstruction also requires modeling appearance, whereas
fitting sparse anatomical labels directly constrains geometry without
reconstructing the underlying image
intensities~\cite{chang2022deeprecon,muffoletto2026neural}.

DMCVR employs morphology-guided diffusion for CMR volume
reconstruction~\cite{he2023dmcvr}, whereas CardiacFlow combines
flow-based augmentation with simulated partial--complete
supervision~\cite{ma2025cardiacflow}. Such supervision provides known
source associations during training, while corresponding dense
anatomies remain unknown for real inputs from independent cohorts.
Multi-atlas methods use similarity-based reference
fusion~\cite{zhuang2016multi}, but unequal anatomical coverage
complicates correspondence between sparse inputs and complete references.

DMCVR conditions its denoising model on separate global semantic and local
morphology representations, using a segmentation encoder to supply
anatomical guidance~\cite{he2023dmcvr}.
This emphasizes the information used to control reconstruction, while the
clinically unpaired setting additionally requires deciding which complete
anatomies should guide a particular sparse observation.

CardiacFlow separates noise-to-shape augmentation from a label-completion
U-Net trained on simulated views of real and generated dense
shapes~\cite{ma2025cardiacflow}.
These choices address reconstruction and prior learning, while our
formulation explicitly selects dense candidate endpoints through
sparse-domain retrieval before conditional transport.

Atlas-based segmentation provides another explicit association between a
target and previously observed anatomy. In multi-modality multi-atlas
segmentation, MRI and CT references are registered to the target and
weighted using both global similarity and multi-scale local
patches~\cite{zhuang2016multi}.
This transfers anatomical information through spatial correspondence and
label fusion. Our memory instead links each simulated sparse acquisition
to its source dense encoding. Comparing a query with sparse keys identifies
relevant observations, while aggregation acts on the linked dense values.
The weights therefore define a latent endpoint rather than a fused label
map, allowing endpoint construction and transport to be learned in
separate stages.

\subsection{Unpaired Translation and Conditional Transport}
Cycle consistency, structural constraints, and contrastive learning
encourage content preservation in unpaired
translation~\cite{zhu2017unpaired,iacono2024structure,wang2023dc,park2020contrastive}.
Specifically, CycleGAN regularizes a forward mapping through reconstruction
by an inverse mapping, whereas CUT aligns corresponding input and output
patches in a learned feature space~\cite{zhu2017unpaired,park2020contrastive}.
CUT uses multilayer patch features and draws negative samples from within
the input image~\cite{park2020contrastive}.
However, cycle consistency favors approximate reversibility, whereas
one sparse observation may admit multiple dense completions.
Preserving observed labels also does not uniquely specify unobserved
anatomy. Diffusion models support conditional
generation~\cite{ho2020denoising,chen2024contourdiff,xia2024diffusion,zhang2025diffusion,luo2024target},
but dense-domain distribution matching alone does not establish
compatibility with an individual sparse observation.

Flow matching learns velocity fields along prescribed
paths~\cite{lipman2022flow,liu2022flow}, with applications to medical
synthesis and conditional
generation~\cite{yazdani2025flow,hadzic2025flow,corso2025composing,rohbeck2025modeling}.
Rectified flow starts from an initial coupling and learns an ODE using
velocities defined along linear interpolations~\cite{liu2022flow}.
This coupling specifies the sample associations used to construct the
regression targets.
For unpaired queries, OKB-CFM makes anatomical association explicit before
transport by comparing sparse queries with simulated sparse keys and
weighting their linked dense values.
Diffusion--flow alignment~\cite{schusterbauer2025diff2flow} and
Schr\"odinger bridges or related
formulations~\cite{de2021diffusion,hong2025trajectory,zhang2025modeling,dong2025cunsb,kim2025unified}
further develop generative transport. Nevertheless, distribution-level
objectives do not by themselves determine query-specific anatomical
correspondence, and independently trained domain encoders lack
guaranteed semantic alignment. Establishing an observation-dependent
source--target association therefore remains a central challenge
in clinically unpaired cardiac completion.

\section{Method}
\label{sec:method}

\subsection{Problem and Domain-Specific Representations}

The clinically unpaired sparse and dense segmentation domains,
$\mathcal X$ and $\mathcal Y$, share five foreground labels:
left-ventricular cavity (LV), left-ventricular myocardium (MYO),
right-ventricular cavity (RV), left atrium (LA), and right atrium (RA).
Both domains use six-channel one-hot tensors, including background,
on a $160\times160\times192$ grid. Given a sparse segmentation
$\mathbf x\in\mathcal X$, we seek a dense completion compatible with
its observed anatomy, without patient-matched clinical pairs.
Unacquired sparse locations are encoded as background, without a
separate observation-mask channel.

In Stage~1, we independently train one variational autoencoder (VAE)
for each domain. Let $r\in\{\mathrm{spa},\mathrm{den}\}$, with
$\mathbf u_{\mathrm{spa}}=\mathbf x$ and
$\mathbf u_{\mathrm{den}}=\mathbf y\in\mathcal Y$.
The encoder $\mathcal E_r$, parameterized by $\phi_r$, predicts the
mean and diagonal covariance of an approximate Gaussian posterior.
A latent tensor is sampled through reparameterization and decoded
to reconstruct the input:
\begin{equation}
\label{eq:domain_vae}
\begin{aligned}
q_{\phi_r}(\mathbf z_r\mid\mathbf u_r)
&=
\mathcal N\!\left(
\boldsymbol\mu_r(\mathbf u_r),
\operatorname{Diag}\!\left(
\boldsymbol\sigma_r^2(\mathbf u_r)
\right)
\right),
\\
\mathbf z_r
&=
\boldsymbol\mu_r(\mathbf u_r)
+
\boldsymbol\sigma_r(\mathbf u_r)
\odot\boldsymbol\epsilon,
\\
\hat{\mathbf u}_r
&=
\mathcal D_r(\mathbf z_r).
\end{aligned}
\end{equation}
Here, $\boldsymbol\mu_r(\mathbf u_r)$ and
$\boldsymbol\sigma_r(\mathbf u_r)$ denote the posterior mean and
element-wise standard deviation, respectively, and $\odot$ denotes
element-wise multiplication. The Gaussian distribution is defined
over the vectorized latent tensor, with
$\operatorname{Diag}$ constructing its diagonal covariance.
The posterior mean, standard deviation, and noise tensor
$\boldsymbol\epsilon$ have the same dimensions as $\mathbf z_r$;
the noise elements are independent standard Gaussian variables.
The domain-specific decoder $\mathcal D_r$ produces the
reconstruction $\hat{\mathbf u}_r$.

Each VAE is optimized independently using voxel-wise $\ell_2$ and
multi-class Dice reconstruction losses, together with KL-divergence
regularization toward a standard Gaussian prior. The reconstruction
terms train each domain-specific representation to recover its input
segmentation labels; no cross-domain alignment is imposed in this stage.
After Stage~1, both VAEs are frozen, and source--target coupling is
introduced through observation-keyed retrieval and matching in Stage~2.
In subsequent stages, $\mathcal E_r(\cdot)$ denotes extraction of a
latent tensor from the frozen VAE representation, whereas
$q_{\phi_r}$ denotes its posterior distribution.

\subsection{Observation-Keyed Matching and Barycentric Endpoint}
\label{sec:retrieval}
\subsubsection{Memory Bank Construction and Observation-keyed Retrieval}
Stage~2 constructs a fixed memory from dense training entries. For each dense annotation $\mathbf y_j$, the acquisition simulator~\cite{ma2025cardiacflow} generates $M=4$ sparse observations $\widetilde{\mathbf x}_{jm}=\mathcal A_{\xi_{jm}}(\mathbf y_j)$, where $m$ indexes the simulated acquisition and $\xi_{jm}$ specifies its plane configuration and perturbations. Using the latent-code extraction notation defined above, each observation and its source are encoded as
\begin{equation}
\big(\mathbf k_{jm},\mathbf v_j,\mathrm{gid}_j\big)
=\big(\mathcal E_{\mathrm{spa}}(\widetilde{\mathbf x}_{jm}),
\mathcal E_{\mathrm{den}}(\mathbf y_j),\mathrm{gid}_j\big),
\label{eq:bank}
\end{equation}
where $\mathbf k_{jm}$ is a sparse-domain key, $\mathbf v_j$ is the dense value shared by its four acquisitions, and $\mathrm{gid}_j$ identifies the dense annotation entry. The stored keys and dense values remain fixed after memory construction. Cached observations are reused as simulated training queries, and acquisition configurations are not resampled during matcher training.

The simulator combines an LA-only two-chamber slab, a central two-slice four-chamber slab retaining all labels, and LV/MYO/RV short-axis planes spanning the myocardial apex-to-valve extent.

To retain global channel statistics and coarse spatial layout, retrieval represents each latent tensor by
\begin{equation}
h(\mathbf z)=\operatorname{norm}\left(
[\operatorname{GAP}(\mathbf z)\|\operatorname{vec}(\operatorname{AAP}_{2\times2\times2}(\mathbf z))]\right),
\label{eq:descriptor}
\end{equation}
where GAP and AAP denote global and adaptive average pooling, respectively; vec denotes vectorization; $\|$ denotes concatenation; and norm denotes $\ell_2$ normalization. With $C_z=256$ latent channels, $h(\mathbf z)\in\mathbb R^{9C_z}$. For a query code $\mathbf z_0=\mathcal E_{\mathrm{spa}}(\mathbf x)$, reused throughout that retrieval operation, sparse-domain cosine similarities are $\rho_{jm}=h(\mathbf z_0)^\top h(\mathbf k_{jm})$. We retain the highest-scoring key for each group and select the top $K=8$ distinct dense entries. The selected key--value pairs are indexed by $\ell$, and their scores are denoted by $\rho_\ell$. Dense values are accessed after ranking. Training, validation, and testing use the same training memory and group-ranking rule, without inserting an additional source candidate for simulated queries.

\subsubsection{Matching Network and Barycentric Endpoint}
\label{sec:matching}
For each retrieved query--key pair, the matching network uses the interaction feature
\begin{equation}
\begin{split}
\boldsymbol\phi_\ell=[h(\mathbf z_0)\|h(\mathbf k_\ell)
\|\{h(\mathbf z_0)-h(\mathbf k_\ell)\}\\
\|\{h(\mathbf z_0)\odot h(\mathbf k_\ell)\}],
\end{split}
\label{eq:phi}
\end{equation}
where $\odot$ denotes element-wise multiplication. The concatenated feature has 9216 elements. A $9216\to512\to1$ MLP with LayerNorm, GELU, and dropout 0.1 produces a scalar compatibility logit $a_\ell$. The logits define candidate weights and a barycentric endpoint:
\begin{equation}
w_\ell=\frac{\exp(a_\ell/\tau)}{\sum_{i=1}^K\exp(a_i/\tau)},\qquad
\mathbf z_{\mathrm{bary}}=\sum_{\ell=1}^K w_\ell\mathbf v_\ell.
\label{eq:bary}
\end{equation}
Here, $\tau$ is the matching temperature and $i$ is the summation index over retrieved candidates. Each normalized weight applies to an entire dense latent tensor. Weights are predicted from sparse query--key evidence, without directly comparing sparse and dense encodings.

Let $\mathcal B_s$ denote a simulated minibatch of size $B_s$. Query $b$ is supervised by a fixed dense source tensor $\mathbf v_{s,b}$, and its retrieved dense values are indexed by $\ell$. The closest retrieved value defines the classification target $\ell_b^\dagger=\arg\min_\ell\operatorname{mean}_e[(v_{b,\ell,e}-v_{s,b,e})^2]$, where $e$ indexes latent elements. Classification and normalized barycentric regression are
\begin{equation}
\begin{aligned}
\mathcal L_{\mathrm{cls}}&=-\frac1{B_s}\sum_{b=1}^{B_s}\log\!\left[\max(w_{b,\ell_b^\dagger},10^{-8})\right],\\
\mathcal L_{\mathrm{bary}}&=\frac{\operatorname{mean}_{b,e}[(z_{\mathrm{bary},b,e}-v_{s,b,e})^2]}
{\max(\operatorname{mean}_{b,e}[v_{s,b,e}^2],10^{-6})}.
\end{aligned}
\label{eq:supervision}
\end{equation}
Both means in the second term span the simulated minibatch and all latent elements; the denominator is detached. The constants $10^{-8}$ and $10^{-6}$ stabilize the logarithm and normalization, respectively. Dense-source distances provide supervision after retrieval and do not alter candidate ranking.

Real cine-CMR queries lack patient-matched dense targets. For these queries, the retrieval prior $p_\ell=\operatorname{softmax}_\ell(\rho_\ell/\tau_{\mathrm{ret}})$ regularizes the learned weights through $\mathcal L_{\mathrm{prior}}=\operatorname{mean}_bD_{\mathrm{KL}}(\mathbf p_b\|\mathbf w_b)$. Here, $\tau_{\mathrm{ret}}$ is the retrieval-prior temperature, $\mathbf p_b$ and $\mathbf w_b$ collect the candidate probabilities for query $b$, and the mean is over the real-query minibatch. This term encourages agreement with the sparse-domain retrieval evidence without defining a dense reference for a real query.

For simulated queries, let $g(\mathbf z)=\operatorname{norm}(\operatorname{GAP}(\mathbf z))$ be the normalized 256-dimensional channel descriptor, and let $\mathcal V_{\mathcal B_s}$ pool the $B_sK$ retrieved dense candidate entries across the minibatch. Pooled-candidate regularization is
\begin{equation}
\mathcal L_{\mathrm{man}}=\frac1{B_s}\sum_{b=1}^{B_s}\min_{\mathbf v\in\mathcal V_{\mathcal B_s}}
\|g(\mathbf z_{\mathrm{bary},b})-g(\mathbf v)\|_2^2.
\label{eq:man}
\end{equation}
This term acts on simulated barycenters and their pooled channel descriptors. The complete matching objective is
\begin{equation}
\mathcal L_{\mathrm{match}}=\mathcal L_{\mathrm{cls}}+\mathcal L_{\mathrm{bary}}
+0.25\mathcal L_{\mathrm{prior}}+0.1\mathcal L_{\mathrm{man}}.
\label{eq:matchloss}
\end{equation}
We use $\tau=0.25$ during training and validation and $\tau_{\mathrm{ret}}=0.07$. The matcher checkpoint minimizes barycentric-endpoint MSE to the fixed dense source encodings of simulated validation queries, averaging first over latent elements and then over query records. Dense validation targets score checkpoints but do not construct their retrieved endpoints. The selected matcher is frozen before flow training; the inference temperature is specified in Section~\ref{sec:flow}.

\subsection{Endpoint-Conditioned Forward Flow}
\label{sec:flow}
With the VAEs, memory, and matcher fixed, Stage~3 learns a velocity field from the sparse encoding $\mathbf z_0$ conditioned on a dense endpoint $\mathbf c$. Simulated training queries use their source encoding $\mathbf c=\mathbf v_s$, whereas real training queries use the matched barycenter $\mathbf c=\mathbf z_{\mathrm{bary}}$. Validation and test endpoints are retrieved barycenters constructed without paired dense targets.

Transport time $t\sim\mathcal U(0.05,0.95)$ is distinct from cardiac phase. Although the configured noise scale is zero, the implemented variance floor of $10^{-8}$ retains a perturbation with standard deviation $10^{-4}$:
\begin{equation}
\mathbf z_t=(1-t)\mathbf z_0+t\mathbf c+10^{-4}\boldsymbol\eta,
\quad \mathbf u_t=\mathbf c-\mathbf z_0,
\label{eq:path}
\end{equation}
where $\boldsymbol\eta$ has the same dimensions as the latent tensors and independent standard Gaussian elements. Equivalently, its vectorized distribution is $\mathcal N(\mathbf 0,\mathbf I)$, with $\mathbf I$ the identity covariance. For a fixed latent pair and noise draw, the perturbation is constant along the path, so $\mathbf u_t=\partial_t\mathbf z_t$. This path noise is distinct from the VAE reparameterization noise $\boldsymbol\epsilon$. Inference integrates over $[0,1]$ from the extracted sparse code without adding this path perturbation.

 Endpoint conditioning uses only $\operatorname{GAP}(\mathbf c)$: its learned embedding is added to the time embedding and injected through adaptive instance normalization. Let $\Phi^{(L)}_\psi$ denote $L$ explicit Euler steps with $\mathbf c$ held fixed. Training minimizes
\begin{equation}
\begin{aligned}
\mathcal L_{\mathrm{flow}}={}&\mathbb E\operatorname{mean}\big[(\mathbf v_\psi(\mathbf z_t,t,\mathbf c)-\mathbf u_t)^2\big]\\
&+\mathbb E\operatorname{mean}\big[(\Phi^{(L)}_\psi(\mathbf z_0;\mathbf c)-\mathbf c)^2\big].
\end{aligned}
\label{eq:flowloss}
\end{equation}
The inner means span latent elements. The first expectation averages over encoded query--endpoint pairs, sampled times, and path perturbations; the second averages terminal errors over the encoded pairs. Both terms use the concatenated simulated and real minibatch. We use $L=8$ and unit endpoint-loss weight. Only $\psi$ is optimized, with upstream modules frozen and endpoints detached. The terminal penalty encourages, but does not enforce, equality with the prescribed endpoint.

The flow checkpoint minimizes the sum of terminal-code MSE to the fixed source encoding and to the retrieved barycenter on simulated validation queries. Each term uses the same element-then-query averaging as matcher selection. Source encodings enter this validation score only; the endpoints remain retrieval-derived. At inference, the matching temperature is $\tau_{\mathrm{infer}}=0.5$, and eight Euler steps yield
\begin{equation}
\begin{aligned}
\mathbf z^{(l+1)}&=\mathbf z^{(l)}+\tfrac18\mathbf v_\psi(\mathbf z^{(l)},l/8,\mathbf c),\\
\hat{\mathbf y}&=\mathcal D_{\mathrm{den}}(\mathbf z^{(8)}),\qquad \mathbf z^{(0)}=\mathbf z_0.
\end{aligned}
\label{eq:inference}
\end{equation}
Here, $l$ runs from 0 to 7. The query is encoded once and reused throughout retrieval and integration. Primary completion uses $\mathbf c=\mathbf z_{\mathrm{bary}}$; alternative completions use individual retrieved values selected according to the same inference-time matching distribution. For fixed latent tensors $\mathbf z_0$ and $\mathbf c$, Euler integration and dense decoding are deterministic.

\begin{table*}[!t]
\centering
\footnotesize\setlength{\tabcolsep}{3pt}
\caption{Real clinically unpaired evaluation on 327 inputs. Enc-Fr\'{e}chet and Enc-MMD are scaled by $10^4$ and $10^7$, respectively. Observation Dice assesses agreement with acquired labels. Mean Dist. is a directed distance in voxels; the separate OKB-CFM evaluation is described below.}
\label{tab:quantitative-main}
\footnotesize\setlength{\tabcolsep}{3.3pt}
\begin{tabular*}{\textwidth}{@{\extracolsep{\fill}}l|cc|ccccc|c}
\toprule
Method & Enc-Fr\'{e}chet $\downarrow$ & Enc-MMD $\downarrow$
& LV Dice$_{\mathrm{obs}}$ $\uparrow$ & MYO $\uparrow$
& RV $\uparrow$ & LA $\uparrow$ & RA $\uparrow$
& Mean Dist. (voxels) $\downarrow$ \\
\midrule
CycleGAN
      & $16.45\pm3.21$
      & $64.99\pm3.64$
      & ${0.848\pm0.034}$
      & ${0.645\pm0.089}$
      & ${0.895\pm0.032}$
      & $0.743\pm0.053$
      & ${0.735\pm0.094}$
      & ${0.882\pm0.287}$ \\
CUT
      & ${14.96\pm4.11}$
      & ${34.84\pm1.51}$
      & $0.833\pm0.034$
      & $0.616\pm0.087$
      & $0.841\pm0.043$
      & ${0.785\pm0.069}$
      & $0.597\pm0.099$
      & $0.991\pm0.332$ \\
EGSDE
      & $34.23\pm4.19$
      & $218.00\pm41.69$
      & $0.652\pm0.139$
      & $0.351\pm0.103$
      & $0.377\pm0.159$
      & $0.537\pm0.198$
      & $0.358\pm0.265$
      & $4.574\pm2.579$ \\
OKB-CFM
      & ${7.93\pm2.90}$
      & ${2.02\pm1.01}$
      & ${0.852\pm0.035}$
      & ${0.651\pm0.080}$
      & ${0.898\pm0.030}$
      & ${0.806\pm0.083}$
      & ${0.759\pm0.090}$
      & $0.942\pm0.447$\textsuperscript{a} \\
\bottomrule
\end{tabular*}
\par\vspace{2pt}
\begin{minipage}{\textwidth}\footnotesize
\textsuperscript{a}OKB-CFM Mean Dist. is obtained from a separate
evaluation on the same 327 inputs using original sparse foreground
labels. Directed same-class distances are averaged over the five
structures within each input; the table reports the mean and population
SD across inputs.
\end{minipage}

\end{table*}
\begin{table*}[!t]
\centering
\footnotesize\setlength{\tabcolsep}{3pt}
\caption{Simulated evaluation on 34 files. (A) Full-volume scores use original annotations for the baselines and dense-VAE reconstructions for OKB-CFM; these groups do not support a common-reference ranking. Distance units are given in the headers. (B) Slice-back consistency with the simulated input.}
\label{tab:dice_pair}
\label{tab:quantitative-surface}
\label{tab:slice-back}
{\footnotesize
\textbf{(A) Full-volume scores by reference construction}\par\vspace{2pt}
\begin{tabular*}{\textwidth}{@{\extracolsep{\fill}}l|ccccc|ccc}
\toprule
Method & LV Dice $\uparrow$ & MYO Dice $\uparrow$ & RV Dice $\uparrow$ &
LA Dice $\uparrow$ & RA Dice $\uparrow$ & CD (voxel$^2$) $\downarrow$ &
HD95 (voxels) $\downarrow$ & ASSD (voxels) $\downarrow$ \\
\midrule
\multicolumn{9}{l}{\textit{Reference: original dense annotations}} \\
CycleGAN & $0.634 \pm 0.039$ & $0.335 \pm 0.052$ & $0.662 \pm 0.059$ & $0.519 \pm 0.090$ & $0.542 \pm 0.070$ & $135.12 \pm 42.87$ & $15.21 \pm 3.59$ & $6.76 \pm 0.92$ \\
CUT & $0.580 \pm 0.029$ & $0.275 \pm 0.042$ & $0.580 \pm 0.055$ & $0.556 \pm 0.093$ & $0.549 \pm 0.059$ & $93.41 \pm 101.43$ & $16.75 \pm 7.39$ & $6.93 \pm 1.62$ \\
EGSDE & $0.638 \pm 0.106$ & $0.352 \pm 0.075$ & $0.396 \pm 0.138$ & $0.580 \pm 0.153$ & $0.487 \pm 0.202$ & $239.97 \pm 93.76$ & $35.64 \pm 5.96$ & $9.58 \pm 2.55$ \\
\midrule
\multicolumn{9}{l}{\textit{Reference: OKB-CFM dense-VAE reconstructions}} \\
OKB-CFM & $0.941 \pm 0.036$ & $0.865 \pm 0.070$ & $0.893 \pm 0.075$ & $0.872 \pm 0.122$ & $0.875 \pm 0.067$ & $6.15 \pm 3.87$ & $3.88 \pm 2.57$ & $1.50 \pm 0.90$ \\
\bottomrule
\end{tabular*}
\vspace{2pt}\par
\textbf{(B) Slice-back consistency}\par\vspace{2pt}
\begin{tabular}{l|ccccc}
\toprule
Method & LV $\uparrow$ & MYO $\uparrow$ & RV $\uparrow$ & LA $\uparrow$ & RA $\uparrow$ \\
\midrule
CycleGAN & $0.708 \pm 0.030$ & $0.442 \pm 0.043$ & $0.788 \pm 0.034$ & $0.675 \pm 0.114$ & $0.783 \pm 0.084$ \\
CUT & $0.664 \pm 0.036$ & $0.380 \pm 0.060$ & $0.683 \pm 0.053$ & $0.785 \pm 0.126$ & $0.653 \pm 0.093$ \\
EGSDE & $0.686 \pm 0.087$ & $0.422 \pm 0.078$ & $0.464 \pm 0.159$ & $0.621 \pm 0.156$ & $0.501 \pm 0.226$ \\
OKB-CFM (ours) & $0.899 \pm 0.057$ & $0.795 \pm 0.106$ & $0.908 \pm 0.058$ & $0.913 \pm 0.105$ & $0.902 \pm 0.062$ \\
\bottomrule
\end{tabular}
}

\end{table*}
\begin{figure*}[!t]
\centering
\includegraphics[width=\textwidth]{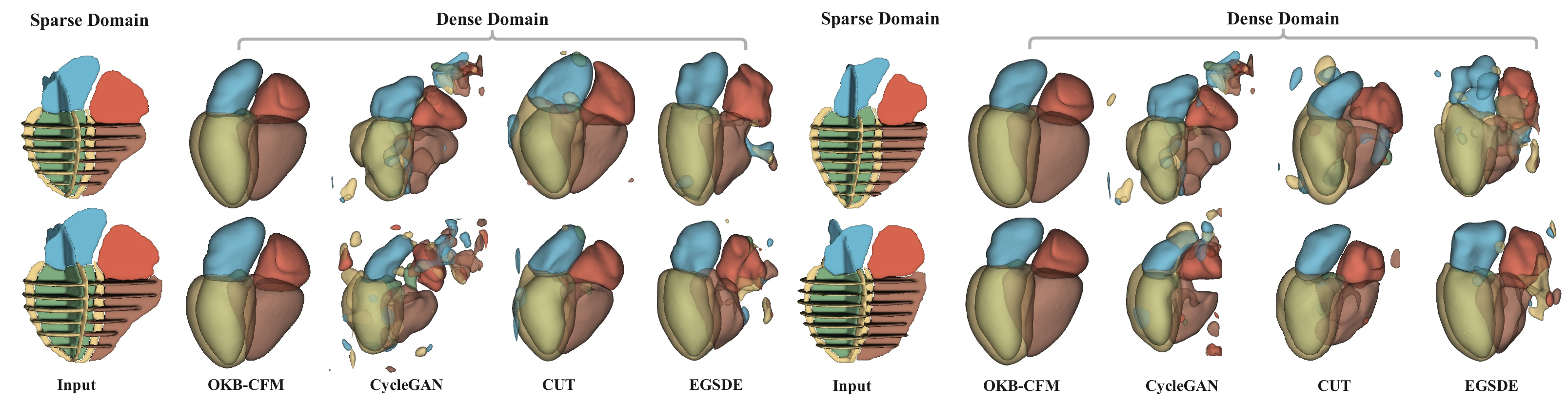}
\caption{Qualitative comparison on four real cine-CMR inputs: sparse observations and completions from OKB-CFM, CycleGAN, CUT, and EGSDE. These clinically unpaired cases have no patient-matched dense references; the views illustrate reconstructed morphology and its relation to the available observations.}
\label{fig:real}
\end{figure*}
\begin{figure}[!t]
\centering
\includegraphics[width=\linewidth]{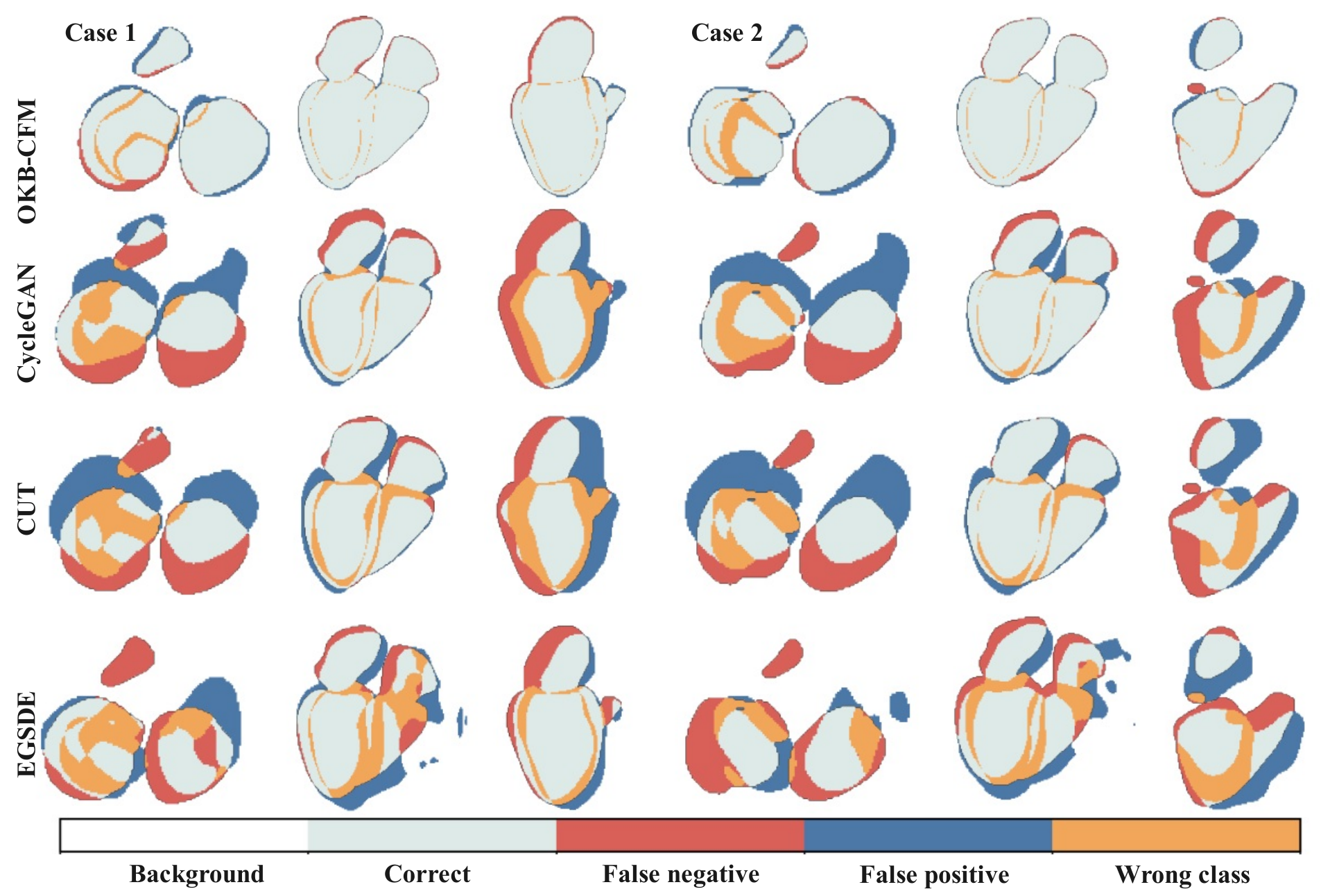}
\caption{Categorical reconstruction errors for two simulated cases. All displayed methods use the same fixed OKB-CFM dense-VAE reconstruction as the reference within each case. This common-reference visualization differs from the original-annotation baseline evaluation in Table~\ref{tab:dice_pair}(A).}
\label{fig:paired}
\end{figure}
\begin{figure*}[!t]
\centering
\includegraphics[width=0.9\textwidth]{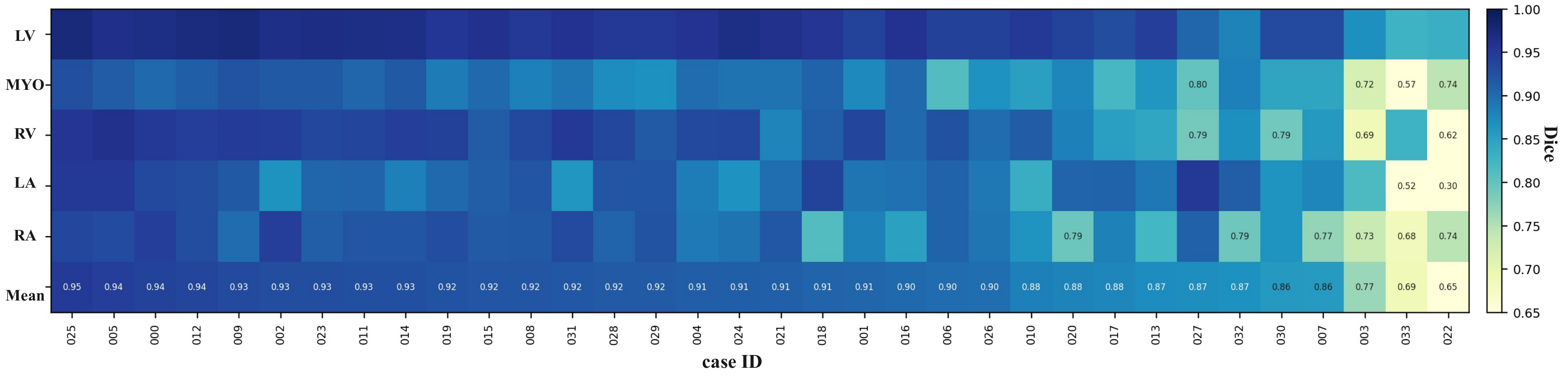}
\caption{Case-wise and structure-wise reconstruction Dice, ordered by mean Dice. Each case is evaluated against its fixed dense-VAE reconstruction. The heatmap complements the cohort averages in Table~\ref{tab:dice_pair}(A) by showing variation across cases and anatomical structures.}
\label{fig:dice-heatmap}
\end{figure*}
\begin{table*}[!t]
\centering
\footnotesize\setlength{\tabcolsep}{3pt}
\caption{Transport and selected training-objective ablations. Real feature discrepancies are scaled by $10^4$ and $10^7$; paired Dice uses dense-VAE references. (A) Direct barycentric decoding with upstream modules and the dense decoder fixed, without flow or retraining. (B) Variants omitting the specified training terms. The full-model row is shared by both panels.}
\label{tab:ablation-components}
\label{tab:ablation-loss-matcher}
{\begin{tabular*}{\textwidth}{@{\extracolsep{\fill}}l|cc|ccccc}
\toprule
Variant & Enc-Fr\'{e}chet $\downarrow$ & Enc-MMD $\downarrow$ &
LV Dice $\uparrow$ & MYO $\uparrow$ & RV $\uparrow$ & LA $\uparrow$ & RA $\uparrow$ \\
\midrule
Full OKB-CFM & $7.93$ & $2.02$ & $0.941$ & $0.865$ & $0.893$ & $0.872$ & $0.875$ \\
\midrule
\multicolumn{8}{l}{\textbf{(A) Finite-step transport versus direct decoding}} \\
Direct barycentric decoding (no flow) & $7.07$ & $1.29$ & $0.895$ & $0.773$ & $0.810$ & $0.805$ & $0.784$ \\
\midrule
\multicolumn{8}{l}{\textbf{(B) Removed training terms}} \\
Without barycentric regression & $10.92$ & $3.68$ & $0.902$ & $0.791$ & $0.833$ & $0.809$ & $0.801$ \\
Without candidate classification & $9.21$ & $2.91$ & $0.923$ & $0.831$ & $0.862$ & $0.840$ & $0.842$ \\
Without real-query retrieval prior & $11.48$ & $4.73$ & $0.916$ & $0.816$ & $0.848$ & $0.824$ & $0.817$ \\
Without endpoint consistency & $9.87$ & $3.15$ & $0.913$ & $0.806$ & $0.842$ & $0.818$ & $0.825$ \\
\bottomrule
\end{tabular*}}
\end{table*}
\begin{table*}[!t]
\centering
\footnotesize\setlength{\tabcolsep}{3pt}
\caption{Retrieval and inference-time candidate count. (A) Reference-selected best-candidate scores against original annotations from 30 files in 29 filename groups, using the fixed 102-entry training memory. Oracle Dice and CD select independently; CD and its gap to the full memory are in voxel$^2$. (B) Inference-time $K$ sweep with networks trained at $K=8$. Real feature metrics are scaled by $10^4$/$10^7$; paired Dice uses 34 dense-VAE references.}
\label{tab:retrieval_oracle}
\label{tab:retrieval_summary}
\label{tab:sensitivity-k}
{\footnotesize
\begin{minipage}[t]{0.465\textwidth}
\centering
\textbf{(A) Retrieval oracle and bank coverage}\par\vspace{2pt}
\setlength{\tabcolsep}{3pt}
\begin{tabular*}{\linewidth}{@{\extracolsep{\fill}}lccc@{}}
\toprule
Candidate set & Oracle Dice $\uparrow$ & Oracle CD $\downarrow$ & CD gap $\downarrow$\\
\midrule
Top-1 & $0.7509$ & $49.134$ & $35.188$\\
Top-2 & $0.7671$ & $22.750$ & $8.805$\\
Top-4 & $0.7847$ & $20.019$ & $6.073$\\
Top-8 & $0.7919$ & $17.420$ & $3.475$\\
Top-16 & $0.7962$ & $16.812$ & $2.866$\\
Full memory & $0.8039$ & $13.946$ & $0.000$\\
\bottomrule
\end{tabular*}
\end{minipage}\hfill
\begin{minipage}[t]{0.51\textwidth}
\centering
\textbf{(B) Inference-time candidate count}\par\vspace{2pt}
\setlength{\tabcolsep}{2pt}
\begin{tabular*}{\linewidth}{@{\extracolsep{\fill}}lrrrrrr@{}}
\toprule
Metric / $K$ & 1 & 2 & 4 & 8 & 16 & 32\\
\midrule
Enc-Fr\'{e}chet $\downarrow$ & 10.84 & 9.66 & 8.41 & 7.93 & 7.96 & 8.12\\
Enc-MMD $\downarrow$ & 3.42 & 2.91 & 2.34 & 2.02 & 2.10 & 2.24\\
LV Dice $\uparrow$ & .861 & .899 & .928 & .941 & .939 & .936\\
MYO Dice $\uparrow$ & .713 & .786 & .846 & .865 & .868 & .860\\
RV Dice $\uparrow$ & .761 & .827 & .875 & .893 & .890 & .886\\
LA Dice $\uparrow$ & .741 & .806 & .851 & .872 & .870 & .866\\
RA Dice $\uparrow$ & .720 & .798 & .855 & .875 & .878 & .873\\
\bottomrule
\end{tabular*}
\end{minipage}
}
\end{table*}
\begin{table}[!t]
\centering
\footnotesize\setlength{\tabcolsep}{3pt}
\caption{Effects of endpoint weighting and conditional transport.
(A) Fixed top-8 candidates and direct decoding without flow.
Real metrics use 327 inputs with original sparse foreground support;
paired Dice uses 34 dense-VAE references.
(B) Regional Dice against original annotations from 23 files in
22 filename groups. Regions follow the geometric acquisition-plane
mask. Gain denotes the absolute Dice increase over direct barycenter
decoding, in percentage points (pp).}
\label{tab:endpoint_controls}
\footnotesize
\setlength{\tabcolsep}{2.5pt}

\textbf{(A) Endpoint weighting without flow}\par\vspace{2pt}
\begin{tabular*}{\linewidth}{@{\extracolsep{\fill}}lccc@{}}
\toprule
Weighting
& \shortstack{Real slice-back\\Dice $\uparrow$}
& \shortstack{Real distance\\(voxels) $\downarrow$}
& \shortstack{Paired mean\\Dice $\uparrow$}\\
\midrule
Uniform & \textbf{0.7538} & \textbf{0.9660} & 0.8140\\
Cosine  & 0.7510 & 0.9807 & \textbf{0.8169}\\
Learned & 0.7532 & 0.9695 & 0.8135\\
\bottomrule
\end{tabular*}

\par\vspace{2pt}
\textbf{(B) Regional effect of transport}\par\vspace{2pt}
\begin{tabular*}{\linewidth}{@{\extracolsep{\fill}}lccc@{}}
\toprule
Region / structure
& \shortstack{Direct\\barycenter}
& OKB-CFM
& \shortstack{Gain\\(pp)}\\
\midrule
Within planes (mean)
& 0.8256 & \textbf{0.8980} & +7.24\\
Outside planes (mean)
& 0.8211 & \textbf{0.8891} & +6.80\\
\midrule
LV outside planes
& 0.8901 & \textbf{0.9352} & +4.51\\
MYO outside planes
& 0.7673 & \textbf{0.8551} & +8.78\\
RV outside planes
& 0.8170 & \textbf{0.8951} & +7.81\\
LA outside planes
& 0.8268 & \textbf{0.8872} & +6.04\\
RA outside planes
& 0.8044 & \textbf{0.8730} & +6.86\\
\bottomrule
\end{tabular*}
\end{table}
\begin{table}[!t]
\centering
\footnotesize\setlength{\tabcolsep}{3pt}
\caption{Candidate completions from a fixed top-8 set. (A) Original references from 30 files in 29 filename groups, with nested sampling with replacement using seed 12345 plus the case index. (B) Observation Dice means $\pm$ population SD on original sparse foreground support for 327 real inputs and 29 simulated filename groups. Minima are over the eight candidates within each file. Distinctness compares exact full-volume hard labels.}
\label{tab:multi_completion_corrected}
\label{tab:sensitivity_N}
\footnotesize
\textbf{(A) Sample quality and endpoint visitation}\par\vspace{2pt}
\setlength{\tabcolsep}{2.5pt}
\begin{tabular*}{\linewidth}{@{\extracolsep{\fill}}cccc@{}}
\toprule
$N$ & \shortstack{Best-of-$N$\\Dice} & \shortstack{Mean\\Dice} & \shortstack{Expected distinct\\endpoints}\\
\midrule
1 & 0.87185 & 0.87185 & 1.000\\
4 & 0.87233 & 0.87195 & 2.999\\
8 & 0.87250 & 0.87198 & 4.547\\
16 & 0.87265 & 0.87198 & 6.119\\
32 & 0.87274 & 0.87201 & 7.255\\
\bottomrule
\end{tabular*}
\par\vspace{2pt}
\textbf{(B) Observation agreement and distinctness}\par\vspace{2pt}
\begin{tabular*}{\linewidth}{@{\extracolsep{\fill}}lcc@{}}
\toprule
Quantity & Real & Simulated\\
\midrule
Weighted mean Dice & $.75647\pm.05726$ & $.87283\pm.06308$\\
Case-wise min. Dice & $.75537\pm.05723$ & $.87195\pm.06305$\\
Distinct outputs & 8 & 8\\
\bottomrule
\end{tabular*}

\end{table}
\begin{figure}[!t]
\centering
\includegraphics[width=0.9\linewidth]{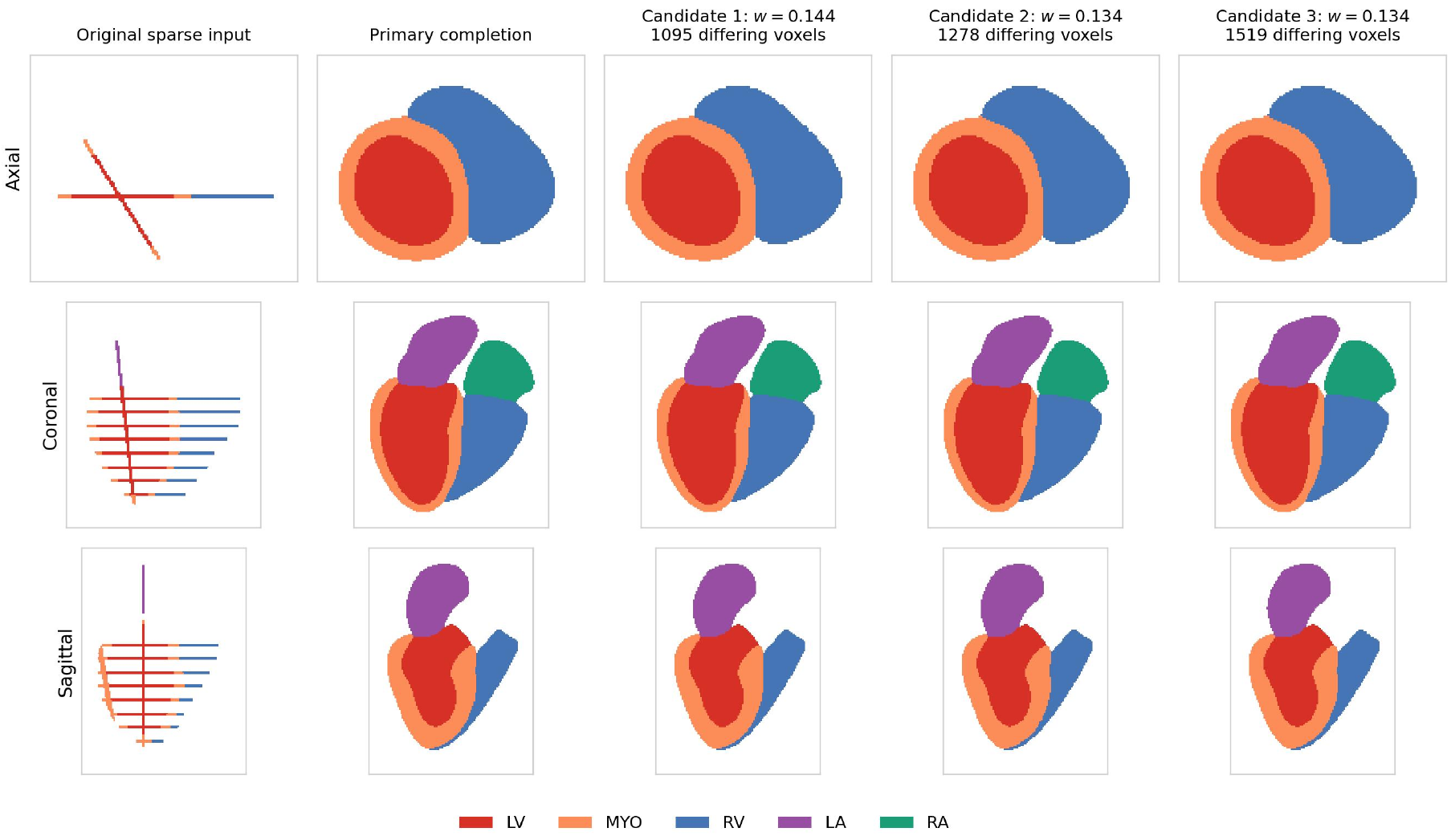}
\caption{Candidate-conditioned completions. Sparse input, primary completion, and the three highest-weight exact-distinct candidate outputs for preselected real entry 0. Weights refer to the full candidate distribution; differing-voxel counts concern complete volumes and need not occur in the displayed slices.}
\label{fig:candidate-outputs-real}
\end{figure}
\begin{figure*}[!t]
\centering
\includegraphics[width=0.85\textwidth]{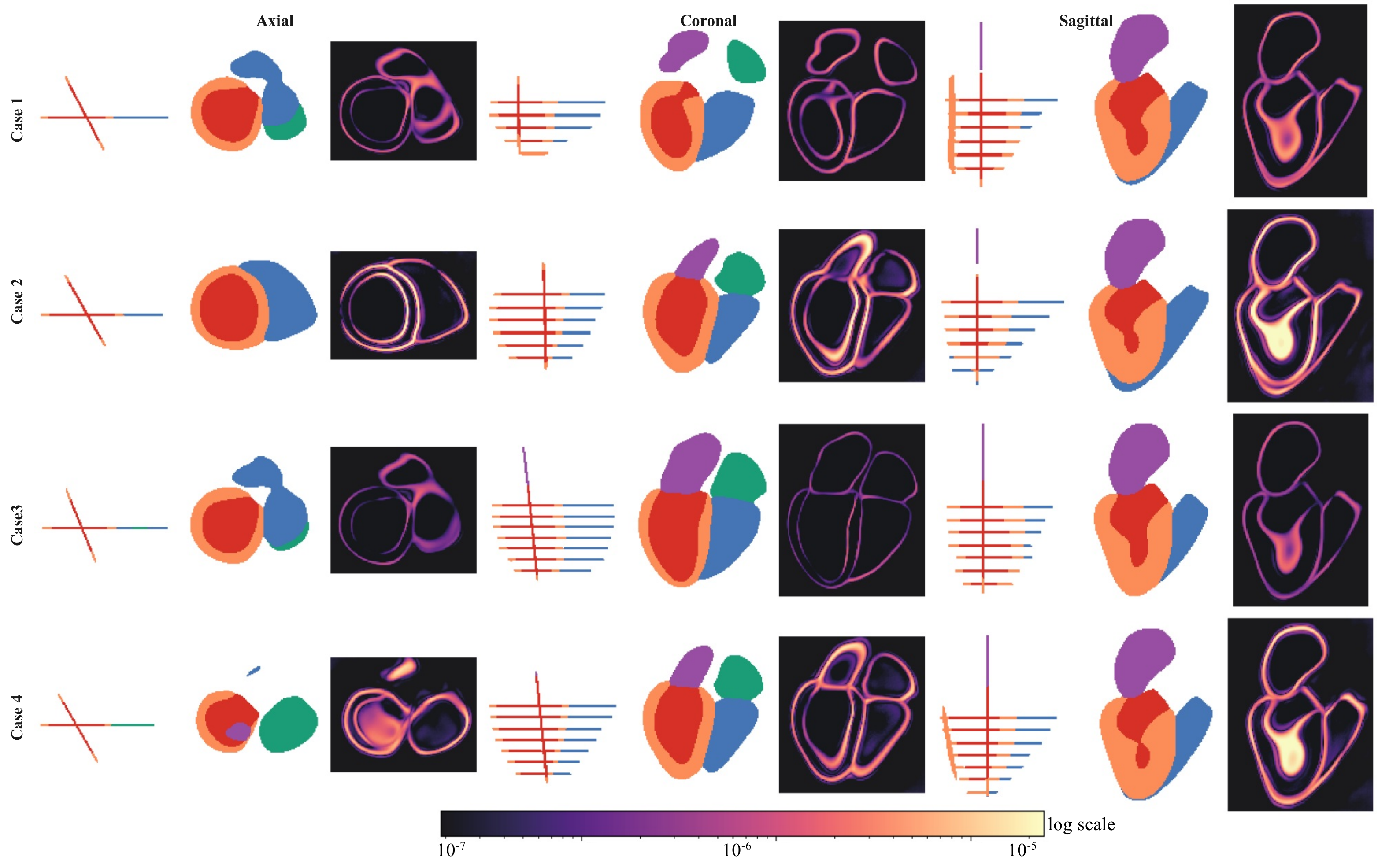}
\caption{Matching-weighted foreground-probability variance for four real cases, using a shared $10^{-7}$--$10^{-5}$ logarithmic scale. The panels do not imply case-wise correspondence. The maps describe variation across the evaluated candidate-conditioned outputs, not calibrated anatomical uncertainty.}
\label{fig:predictive-dispersion}
\end{figure*}
\begin{figure}[!t]
\centering
\includegraphics[width=0.85\linewidth]{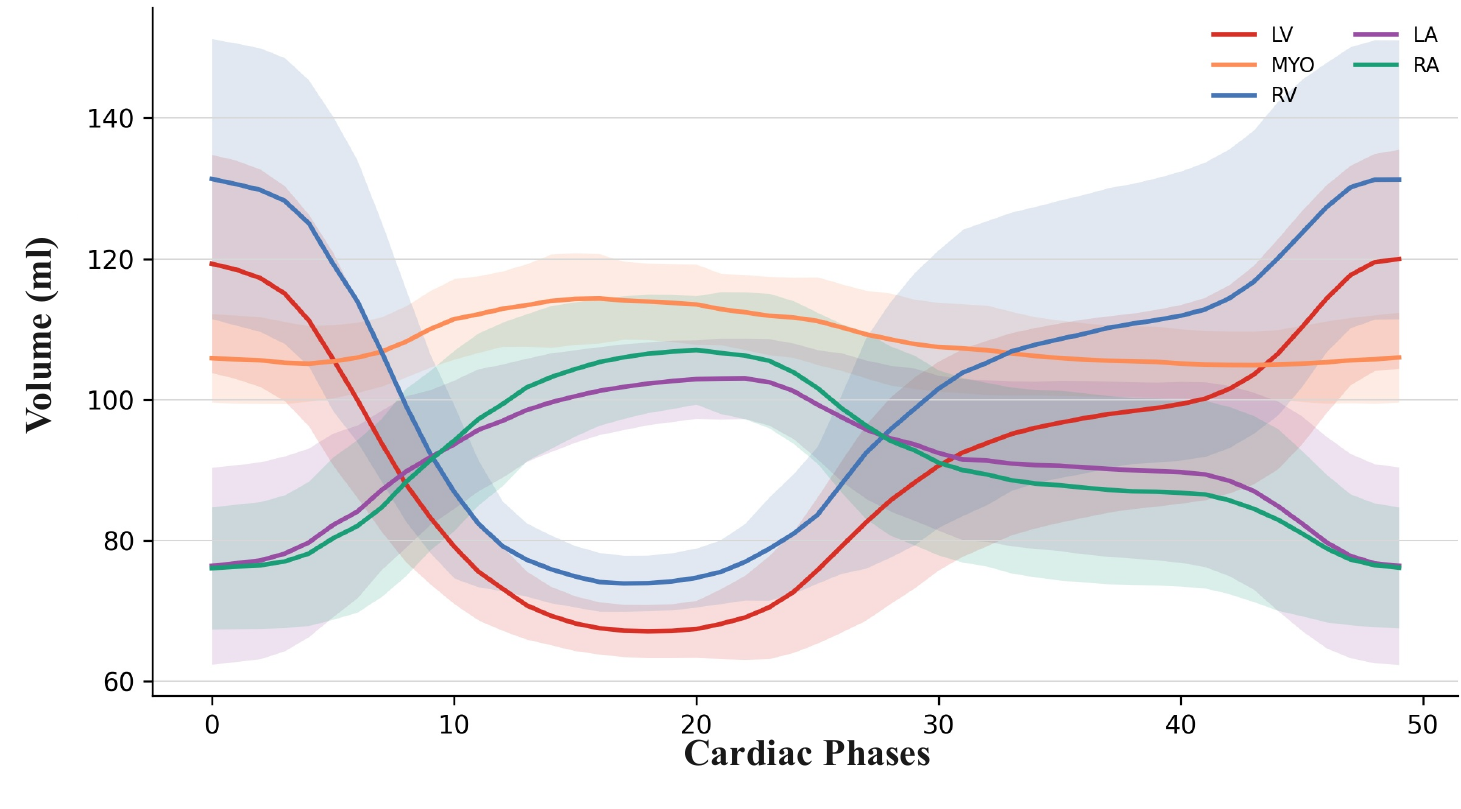}
\caption{Frame-wise cine application. Population-level cardiac-volume trajectories (ml).}
\label{fig:temporal-volume}
\end{figure}
\begin{figure*}[!t]
\centering
\includegraphics[width=0.85\textwidth]{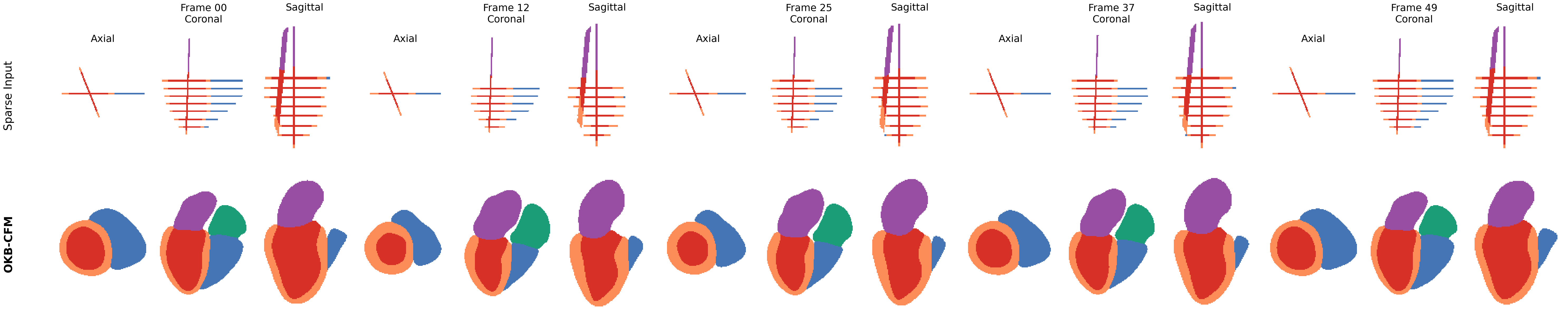}
\caption{Sparse observations and completions at frames 0, 12, 25, 37, and 49 of one 50-frame sequence, using fixed axial, coronal, and sagittal positions and a common crop. Each frame is processed independently.}
\label{fig:cine-completion}
\end{figure*}

\section{Experimental Settings}
\label{sec:experimental_settings}
\subsection{Data and Partitioning}
The sparse domain contains 1,629 UK Biobank cine-CMR entries with five cardiac labels obtained using a pretrained nnU-Net~\cite{dillon2025open}. Phase-matched short- and long-axis labels are mapped to patient coordinates using DICOM metadata and fused into sparse 3D stacks~\cite{bai2018automated}. Real and simulated sparse observations share an anatomical orientation and cardiac-centered coordinate frame. The dense domain contains 170 whole-heart entries~\cite{zhuang2018multivariate}, affinely aligned to a population cardiac atlas using FLIRT~\cite{zhuang2016multi,fischer2003flirt}. Both domains use 1.25-mm isotropic spacing and the input grid specified in Section~\ref{sec:method}.

Independent permutations with seed 12345 produce 977/325/327 sparse and 102/34/34 dense training/validation/test entries. Sparse indices identify folders, using the first frame only, while dense indices identify annotation files. Repeated dense annotations were not grouped before splitting; filename-prefix overlaps comprise one training--validation group and four training--test groups. Because these prefixes have not been verified as patient identifiers, patient-level separation is not established. The fixed memory contains 408 keys from the 102 dense training entries. Simulated validation and test observations follow the corresponding dense-file partitions.

\subsection{Optimization and Baselines}
All three stages use AdamW with learning rate $10^{-4}$ and maximum epoch counts of 200, 30, and 100 for the VAEs, matcher, and flow network, respectively.  A single-stack inference takes approximately 17.23 seconds on an NVIDIA A6000 GPU.

We compare with latent-space CycleGAN~\cite{zhu2017unpaired}, volumetric PatchNCE-based CUT~\cite{park2020contrastive}, and energy-guided diffusion EGSDE~\cite{zhao2022egsde}. These methods use separate generation encoders and decoders; all outputs are assessed with the same frozen evaluation encoder. The available legacy CardiacFlow checkpoint overlaps the reference cohort and was not retrained on the reported split, so it is excluded from the primary comparison.


\subsection{Metrics and Analysis Protocols}

Real unpaired evaluation assesses distributional agreement using
Enc-Fr\'{e}chet and Enc-MMD in a shared 256-dimensional feature space
obtained by global average pooling from an encoder trained only on
dense training data. All 327 completions and 34 dense test volumes are
included.  Enc-Fr\'{e}chet and Enc-MMD are
scaled by $10^4$ and $10^7$, respectively.

Observation consistency is measured by slice-back Dice and Mean Dist.
Slice-back Dice evaluates label agreement at acquired locations,
whereas Mean Dist. measures the directed nearest-neighbor distance
from observed to generated foreground within each anatomical class.
Distances are reported in voxels and averaged across the five
structures and then across inputs. The separate OKB-CFM Mean Dist.
evaluation and endpoint/candidate analyses use the original sparse
foreground.

Simulated reconstruction is evaluated using Dice, squared Chamfer
distance (CD), HD95, and average symmetric surface distance (ASSD).
 Results are
summarized by means and population SDs over 34 test annotations.
Baseline scores use original annotations, whereas primary OKB-CFM
scores use dense-VAE-reconstructed references; these results therefore
do not constitute a common-reference ranking.

\section{Results}
\label{sec:results}
\subsection{Quantitative Evaluation}
\subsubsection{Real Clinically Unpaired Evaluation}
Table~\ref{tab:quantitative-main} reports cohort-level feature agreement and consistency with acquired labels. OKB-CFM has lower Enc-Fr\'{e}chet and Enc-MMD than the three translation baselines, with reductions of 47.0\% and 94.2\% relative to CUT. Its observation Dice is higher for all five structures, although the LV, MYO, and RV differences from CycleGAN are small and are interpreted descriptively. The separate original-foreground evaluation gives a Mean Dist. of $0.942\pm0.447$ voxels; CycleGAN has a lower reported directed distance. The results therefore support improved feature-distribution and observed-label agreement under the reported protocols.

\subsubsection{Simulated Reconstruction}
Table~\ref{tab:dice_pair}(A) groups full-volume results by reference construction. Against dense-VAE-reconstructed references, OKB-CFM obtains a five-class mean Dice of 0.889, CD of 6.15 voxel$^2$, HD95 of 3.88 voxels, and ASSD of 1.50 voxels. The baseline full-volume scores use original annotations and are therefore not ranked against the OKB-CFM row. Panel B evaluates agreement with the simulated sparse inputs through slice-back Dice, which is higher for OKB-CFM across all five structures.

\subsection{Qualitative Comparison and Case Variation}
Figure~\ref{fig:real} compares four real cine-CMR examples. In the displayed cases, OKB-CFM produces more connected structures and clearer chamber separation, while the translation baselines exhibit fragmentation, missing structures, or local distortion. These examples illustrate morphology and agreement with the available sparse labels; no patient-matched dense reference is available.

Figure~\ref{fig:paired} compares categorical reconstruction errors
in two simulated cases. Compared with CycleGAN, CUT, and EGSDE,
OKB-CFM shows less extensive false-positive and false-negative
regions along chamber contours, while wrong-class predictions
are more localized. These observations indicate closer boundary
agreement and reduced confusion between anatomical labels in
the displayed cases. Residual errors remain in thin myocardial
regions and at some chamber interfaces.

Figure~\ref{fig:dice-heatmap} shows generally high and relatively
consistent LV Dice across cases, with greater variation in the
other structures. MYO achieves a lower cohort Dice than LV and RV
(0.865 versus 0.941 and 0.893), consistent with the myocardial
errors visible in Fig.~\ref{fig:paired}. The lower-scoring cases
also exhibit pronounced reductions in RV and atrial Dice.
These findings show that comparatively preserved LV agreement
can coexist with substantial errors in other structures,
highlighting case-dependent limitations in multi-chamber
reconstruction.

\subsection{Endpoint Construction and Flow Ablations}
\subsubsection{Transport and Training Objectives}
Table~\ref{tab:ablation-components}(A) compares the full model with direct decoding of the same barycenter, keeping the latent codes, memory, matcher, and epoch-200 dense decoder fixed without retraining. Direct decoding yields lower feature-distribution discrepancies (7.07 and 1.29 versus 7.93 and 2.02), whereas finite-step transport improves Dice for every structure under dense-VAE references.

Table~\ref{tab:ablation-components}(B) evaluates four selected training terms. Removing barycentric regression reduces all five paired Dice scores more than removing candidate classification. Removing the real-query retrieval prior produces the largest increase in both feature-distribution discrepancies among the tested objective variants. Removing endpoint consistency also reduces all five paired Dice scores.

\subsubsection{Retrieval and Endpoint Controls}
Table~\ref{tab:retrieval_summary}(A) examines whether sparse-domain retrieval retains anatomically compatible candidates from the fixed memory. Expanding from top-1 to top-8 improves reference-selected best-candidate Dice and CD, although a gap to the full memory remains. Dice and CD select candidates independently. These scores characterize individual candidates available before aggregation and transport; they are not bounds on the decoded barycenter or the flow output.

With the networks trained at $K=8$ held fixed, the inference-time sweep in Table~\ref{tab:retrieval_summary}(B) improves all five paired Dice scores and both feature-distribution metrics from $K=1$ to $8$. Changes at $K=16$ and $32$ are smaller and mixed. These results support $K=8$ as the reported operating point for this memory and model.

Table~\ref{tab:endpoint_controls}(B) demonstrates the contribution
of conditional transport to regional anatomical reconstruction.
Compared with direct barycenter decoding, OKB-CFM increases mean
Dice by 7.24 percentage points within the acquired planes and
6.80 percentage points outside them, evaluated against original
dense annotations. The improvement extends to all five structures
outside the acquired planes, with the largest gains in MYO and RV
(8.78 and 7.81 percentage points, respectively). These results
support the effectiveness of conditional transport in recovering
multi-chamber anatomy beyond directly sampled regions under
simulated acquisitions.

Table~\ref{tab:endpoint_controls}(A) examines endpoint aggregation
using fixed candidates and direct decoding. Uniform, cosine-similarity,
and learned weighting yield numerically close scores, with the
learned-weight output also serving as the no-flow control in
Table~\ref{tab:ablation-components}. The separate terminal-state
analysis reports a mean latent RMSE of 0.02572 and a relative
endpoint deviation of 0.71806, quantifying the difference between
the learned finite-step output and its prescribed barycenter.

\subsection{Candidate-Conditioned Completions}

Table~\ref{tab:multi_completion_corrected}(A) shows that OKB-CFM
maintains stable average reconstruction quality as more endpoints
are sampled from the fixed top-$8$ set. Increasing $N$ from 1 to 32
raises the expected number of distinct endpoints visited from
1.000 to 7.255, while mean sampled Dice remains between 0.87185
and 0.87201. Best-of-$N$ Dice, obtained through reference-based
selection, increases from 0.87185 to 0.87274. Thus, additional
sampling accesses more candidate endpoints while preserving
average reconstruction accuracy.

The candidate-conditioned outputs also exhibit consistent agreement
with the acquired cardiac labels
(Table~\ref{tab:multi_completion_corrected}(B)).
The differences between matching-weighted mean and mean case-wise
minimum slice-back Dice are only 0.00110 for real inputs and
0.00088 for simulated inputs, indicating similar observation
agreement across candidate choices on average.
All eight outputs are distinct under exact full-volume hard-label
comparison. Their matching-weighted pairwise CD of
0.05018 voxel$^2$ indicates that these alternatives remain
geometrically close, providing small shape variations while
maintaining comparable agreement with the sparse observations.

Figure~\ref{fig:candidate-outputs-real} shows the primary completion and three highest-weight exact-distinct candidate outputs for preselected real entry 0. The accompanying probability-variance maps show four real cases with their own anatomical views, using a shared logarithmic scale of $10^{-7}$--$10^{-5}$. These maps localize variation across the evaluated candidate-conditioned outputs and complement the geometric measurements.

\subsection{Frame-Wise Cine Application}
The trained static model is applied independently to successive cine frames (Fig.~\ref{fig:temporal-volume}). The reported population-level LV and RV volume trajectories decrease during early-to-middle phases and subsequently increase; atrial trajectories show approximately opposite trends, while MYO varies less. Five frames from one 50-frame sequence illustrate phase-dependent outputs at fixed anatomical positions and with a common crop.

\section{Discussion and Limitations}

\looseness=-1
OKB-CFM addresses a practical challenge in cardiac imaging:
sparse cine-CMR segmentations and complete cardiac annotations
are often available from independent cohorts without patient-matched
pairs. Observation-keyed retrieval connects the acquired anatomy
to dense shape information through the sparse representation.
The regional evaluation against original annotations shows that
conditional transport improves reconstruction both within and
outside the acquired planes, with improvements across all five
structures outside the planes. These findings support the use
of independently collected anatomical data for completing the
limited spatial coverage of cine imaging. Such completion may
provide additional anatomical context for studying chamber
morphology alongside cardiac function.

\looseness=-1
The component analyses help distinguish the roles of endpoint
construction and transport. The loss ablations support the
contributions of dense-source supervision, real-query retrieval
regularization, and endpoint consistency within the full model.
Transport improves paired reconstruction accuracy, whereas direct
barycenter decoding produces lower feature-distribution
discrepancies. This difference highlights the complementary
information provided by distributional and anatomical metrics.
The fixed-candidate controls yield similar scores across weighting
schemes without flow, leaving the independent benefit of learned
weighting within the complete pipeline unresolved. Terminal
diagnostics also show that the learned output differs from its
prescribed endpoint; the current results do not explain which
of these changes contribute to the anatomical improvements.

Several limitations affect the interpretation of the evaluations.
Real cine-CMR inputs lack patient-matched dense annotations,
so agreement with observed labels and the dense cohort cannot
verify anatomy outside the acquired planes. In addition, the
primary paired comparison uses different reference definitions
for OKB-CFM and the baselines, preventing a direct ranking of
their full-volume accuracy. The baselines also use separate
generation encoders and decoders, so these comparisons evaluate
complete systems and do not isolate the contribution of the
proposed coupling.

Finally, candidate retrieval depends on the anatomies represented
in the fixed memory. Pooled descriptors may overlook local
differences, and latent averaging does not guarantee anatomically
valid shapes. The candidate-conditioned outputs maintain similar
observation agreement but exhibit limited geometric variation;
their variability has not been validated as an estimate of
anatomical uncertainty. Application to successive cine frames
demonstrates completion across cardiac phases, while temporal
consistency and the accuracy of derived functional measurements
remain unevaluated.

\section{Conclusion}

We presented OKB-CFM for multi-chamber cardiac shape completion
from sparse cine-CMR segmentations using clinically unpaired
dense anatomical data. The framework combines sparse-domain
retrieval, barycentric endpoint construction, and conditional
latent transport without assuming semantic alignment between
independently trained VAEs. Real-data evaluation shows lower
feature-distribution discrepancies and higher observed-label
agreement than the evaluated unpaired translation baselines.
Regional analyses against original dense annotations further
show that transport improves reconstruction across all five
cardiac structures outside the acquired planes.
Candidate-conditioned outputs also provide alternative completions
with similar agreement with the sparse observations.

These findings support the use of observation-conditioned
anatomical information from independent cohorts to complete
the limited spatial coverage of cine imaging. By providing
inferred multi-chamber anatomical context beyond the acquired
planes, OKB-CFM offers a basis for investigating chamber
morphology and spatial relationships alongside cardiac function.
This connection between sparse functional imaging and
three-dimensional anatomical modeling provides a direction
for more integrated studies of cardiac structure and function.

\bibliographystyle{IEEEtran}
\bibliography{references}

@inproceedings{ma2025cardiacflow,
  title={CardiacFlow: 3D+ t Four-Chamber Cardiac Shape Completion and Generation via Flow Matching},
  author={Ma, Qiang and Meng, Qingjie and Qiao, Mengyun and Matthews, Paul M and O'Regan, Declan P and Bai, Wenjia},
  booktitle={International Conference on Medical Image Computing and Computer-Assisted Intervention},
  pages={89--99},
  year={2025},
}

@inproceedings{dillon2025open,
  title={An open-source end-to-end pipeline for generating 3D+ t biventricular meshes from cardiac magnetic resonance imaging},
  author={Dillon, Joshua R and Mauger, Charl{\`e}ne and Zhao, Debbie and Deng, Yu and Petersen, Steffen E and McCulloch, Andrew D and Young, Alistair A and Nash, Martyn P},
  booktitle={International Conference on Functional Imaging and Modeling of the Heart},
  pages={372--383},
  year={2025},
  organization={Springer}
}

@article{bai2018automated,
  title={Automated cardiovascular magnetic resonance image analysis with fully convolutional networks},
  author={Bai, Wenjia and Sinclair, Matthew and Tarroni, Giacomo and Oktay, Ozan and Rajchl, Martin and Vaillant, Ghislain and Lee, Aaron M and Aung, Nay and Lukaschuk, Elena and Sanghvi, Mihir M and others},
  journal={Journal of cardiovascular magnetic resonance},
  volume={20},
  number={1},
  pages={65},
  year={2018},
  publisher={Elsevier}
}

@inproceedings{fischer2003flirt,
  title={FLIRT: A flexible image registration toolbox},
  author={Fischer, Bernd and Modersitzki, Jan},
  booktitle={International workshop on biomedical image registration},
  pages={261--270},
  year={2003},
  organization={Springer}
}

@article{zhuang2018multivariate,
  title={Multivariate mixture model for myocardial segmentation combining multi-source images},
  author={Zhuang, Xiahai},
  journal={IEEE transactions on pattern analysis and machine intelligence},
  volume={41},
  number={12},
  pages={2933--2946},
  year={2018},
  publisher={IEEE}
}

@article{zhuang2016multi,
  title={Multi-scale patch and multi-modality atlases for whole heart segmentation of MRI},
  author={Zhuang, Xiahai and Shen, Juan},
  journal={Medical image analysis},
  volume={31},
  pages={77--87},
  year={2016},
  publisher={Elsevier}
}

@inproceedings{zhu2017unpaired,
  title={Unpaired image-to-image translation using cycle-consistent adversarial networks},
  author={Zhu, Jun-Yan and Park, Taesung and Isola, Phillip and Efros, Alexei A},
  booktitle={Proceedings of the IEEE international conference on computer vision},
  pages={2223--2232},
  year={2017}
}

@article{liu2022flow,
  title={Flow straight and fast: Learning to generate and transfer data with rectified flow},
  author={Liu, Xingchao and Gong, Chengyue and Liu, Qiang},
  journal={arXiv :2209.03003},
  year={2022}
}

@article{pandey2025deep,
  title={Deep learning in 3D cardiac reconstruction: a systematic review of methodologies and dataset},
  author={Pandey, Rajendra Kumar and Rathore, Yogesh Kumar},
  journal={Medical \& Biological Engineering \& Computing},
  volume={63},
  number={5},
  pages={1271--1287},
  year={2025},
  publisher={Springer}
}

@inproceedings{chang2022deeprecon,
  title={Deeprecon: Joint 2d cardiac segmentation and 3d volume reconstruction via a structure-specific generative method},
  author={Chang, Qi and Yan, Zhennan and Zhou, Mu and Liu, Di and Sawalha, Khalid and Ye, Meng and Zhangli, Qilong and Kanski, Mikael and Al'Aref, Subhi and Axel, Leon and others},
  booktitle={International Conference on Medical Image Computing and Computer-Assisted Intervention},
  pages={567--577},
  year={2022},
}

@inproceedings{yang2023whole,
  title={Whole-heart reconstruction with explicit topology integrated learning},
  author={Yang, Huilin and Tam, Roger and Tang, Xiaoying},
  booktitle={International conference on medical image computing and computer-assisted intervention},
  pages={106--115},
  year={2023},
}

@article{sander2023reconstruction,
  title={Reconstruction and completion of high-resolution 3D cardiac shapes using anisotropic CMRI segmentations and continuous implicit neural representations},
  author={Sander, J{\"o}rg and de Vos, Bob D and Bruns, Steffen and Planken, Nils and Viergever, Max A and Leiner, Tim and I{\v{s}}gum, Ivana},
  journal={Computers in Biology and Medicine},
  volume={164},
  pages={107266},
  year={2023},
}

@article{muffoletto2026neural,
  title={Neural Implicit Heart Coordinates: 3D cardiac shape reconstruction from sparse segmentations},
  author={Muffoletto, Marica and Hermida, Uxio and Mauger, Charl{\`e}ne and Suinesiaputra, Avan and Xu, Yiyang and Burns, Richard and Pankewitz, Lisa and Mcculloch, Andrew D and Petersen, Steffen E and Rueckert, Daniel and others},
  journal={Medical Image Analysis},
  pages={104052},
  year={2026},
}

@inproceedings{he2023dmcvr,
  title={Dmcvr: Morphology-guided diffusion model for 3d cardiac volume reconstruction},
  author={He, Xiaoxiao and Tan, Chaowei and Han, Ligong and Liu, Bo and Axel, Leon and Li, Kang and Metaxas, Dimitris N},
  booktitle={International conference on medical image computing and computer-assisted intervention},
  pages={132--142},
  year={2023},
}

@inproceedings{iacono2024structure,
  title={Structure preserving cycle-gan for unsupervised medical image domain adaptation},
  author={Iacono, Paolo and Khan, Naimul},
  booktitle={International Conference of the IEEE Engineering in Medicine and Biology Society},
  pages={1--5},
  year={2024}
}

@article{wang2023dc,
  title={DC-cycleGAN: Bidirectional CT-to-MR synthesis from unpaired data},
  author={Wang, Jiayuan and Wu, QM Jonathan and Pourpanah, Farhad},
  journal={Computerized Medical Imaging and Graphics},
  volume={108},
  pages={102249},
  year={2023},
}

@inproceedings{park2020contrastive,
  title={Contrastive learning for unpaired image-to-image translation},
  author={Park, Taesung and Efros, Alexei A and Zhang, Richard and Zhu, Jun-Yan},
  booktitle={European conference on computer vision},
  pages={319--345},
  year={2020},
}

@article{chen2024contourdiff,
  title={ContourDiff: Unpaired Medical Image Translation with Structural Consistency},
  author={Chen, Yuwen and Konz, Nicholas and Gu, Hanxue and Dong, Haoyu and Chen, Yaqian and Li, Lin and Lee, Jisoo and Mazurowski, Maciej A},
  journal={arXiv :2403.10786},
  year={2024}
}

@article{lipman2022flow,
  title={Flow matching for generative modeling},
  author={Lipman, Yaron and Chen, Ricky TQ and Ben-Hamu, Heli and Nickel, Maximilian and Le, Matt},
  journal={arXiv :2210.02747},
  year={2022}
}

@article{de2021diffusion,
  title={Diffusion schr{\"o}dinger bridge with applications to score-based generative modeling},
  author={De Bortoli, Valentin and Thornton, James and Heng, Jeremy and Doucet, Arnaud},
  journal={Advances in neural information processing systems},
  volume={34},
  pages={17695--17709},
  year={2021}
}

@article{xia2024diffusion,
  title={A diffusion model translator for efficient image-to-image translation},
  author={Xia, Mengfei and Zhou, Yu and Yi, Ran and Liu, Yong-Jin and Wang, Wenping},
  journal={IEEE Transactions on Pattern Analysis and Machine Intelligence},
  volume={46},
  number={12},
  pages={10272--10283},
  year={2024},
  publisher={IEEE}
}

@inproceedings{zhang2025diffusion,
  title={Diffusion Bridge Models for 3D Medical Image Translation},
  author={Zhang, Shaorong and Chattopadhyay, Tamoghna and Thomopoulos, Sophia I and Ambite, Jose-Luis and Thompson, Paul M and Ver Steeg, Greg},
  booktitle={2025 47th Annual International Conference of the IEEE Engineering in Medicine and Biology Society (EMBC)},
  pages={1--7},
  year={2025},
  organization={IEEE}
}

@article{luo2024target,
  title={Target-guided diffusion models for unpaired cross-modality medical image translation},
  author={Luo, Yimin and Yang, Qinyu and Liu, Ziyi and Shi, Zenglin and Huang, Weimin and Zheng, Guoyan and Cheng, Jun},
  journal={IEEE Journal of Biomedical and Health Informatics},
  volume={28},
  number={7},
  pages={4062--4071},
  year={2024},
  publisher={IEEE}
}

@article{ho2020denoising,
  title={Denoising diffusion probabilistic models},
  author={Ho, Jonathan and Jain, Ajay and Abbeel, Pieter},
  journal={Advances in neural information processing systems},
  volume={33},
  pages={6840--6851},
  year={2020}
}

@inproceedings{yazdani2025flow,
  title={Flow matching for medical image synthesis: Bridging the gap between speed and quality},
  author={Yazdani, Milad and Medghalchi, Yasamin and Ashrafian, Pooria and Hacihaliloglu, Ilker and Shahriari, Dena},
  booktitle={International Conference on Medical Image Computing and Computer-Assisted Intervention},
  pages={216--226},
  year={2025},
}

@article{hadzic2025flow,
  title={Flow matching-based data synthesis for robust anatomical landmark localization},
  author={Hadzic, Arnela and Bogensperger, Lea and Berghold, Andrea and Urschler, Martin},
  journal={IEEE Journal of Biomedical and Health Informatics},
  year={2025},
  publisher={IEEE}
}

@inproceedings{corso2025composing,
  title={Composing unbalanced flows for flexible docking and relaxation},
  author={Corso, Gabriele and Somnath, Vignesh Ram and Getz, Noah and Barzilay, Regina and Jaakkola, Tommi and Krause, Andreas},
  booktitle={The Thirteenth International Conference on Learning Representations},
  year={2025}
}

@inproceedings{rohbeck2025modeling,
  title={Modeling complex system dynamics with flow matching across time and conditions},
  author={Rohbeck, Martin and De Brouwer, Edward and Bunne, Charlotte and Huetter, Jan-Christian and Biton, Anne and Chen, Kelvin Y and Regev, Aviv and Lopez, Romain},
  booktitle={The Thirteenth International Conference on Learning Representations},
  year={2025}
}

@article{hong2025trajectory,
  title={Trajectory inference with smooth schr$\backslash$" odinger bridges},
  author={Hong, Wanli and Shi, Yuliang and Niles-Weed, Jonathan},
  journal={arXiv preprint arXiv:2503.00530},
  year={2025}
}

@article{zhang2025modeling,
  title={Modeling cell dynamics and interactions with unbalanced mean field schr$\backslash$" odinger bridge},
  author={Zhang, Zhenyi and Wang, Zihan and Sun, Yuhao and Li, Tiejun and Zhou, Peijie},
  journal={arXiv preprint arXiv:2505.11197},
  year={2025}
}

@inproceedings{dong2025cunsb,
  title={Cunsb-rfie: Context-aware unpaired neural schr{\"o}dinger bridge in retinal fundus image enhancement},
  author={Dong, Xuanzhao and Vasa, Vamsi Krishna and Zhu, Wenhui and Qiu, Peijie and Chen, Xiwen and Su, Yi and Xiong, Yujian and Yang, Zhangsihao and Chen, Yanxi and Wang, Yalin},
  booktitle={2025 IEEE/CVF Winter Conference on Applications of Computer Vision (WACV)},
  pages={4502--4511},
  year={2025},
  organization={IEEE}
}

@article{kim2025unified,
  title={A Unified Framework for Diffusion Bridge Problems: Flow Matching and Schr$\backslash$"$\{$o$\}$ dinger Matching into One},
  author={Kim, Minyoung},
  journal={arXiv preprint arXiv:2503.21756},
  year={2025}
}

@inproceedings{schusterbauer2025diff2flow,
  title={Diff2flow: Training flow matching models via diffusion model alignment},
  author={Schusterbauer, Johannes and Gui, Ming and Fundel, Frank and Ommer, Bj{\"o}rn},
  booktitle={Proceedings of the Computer Vision and Pattern Recognition Conference},
  pages={28347--28357},
  year={2025}
}

@article{zhao2022egsde,
  title={Egsde: Unpaired image-to-image translation via energy-guided stochastic differential equations},
  author={Zhao, Min and Bao, Fan and Li, Chongxuan and Zhu, Jun},
  journal={Advances in Neural Information Processing Systems},
  volume={35},
  pages={3609--3623},
  year={2022}
}

@article{zhang2025towards,
  title={Towards cardiac MRI foundation models: Comprehensive visual-tabular representations for whole-heart assessment and beyond},
  author={Zhang, Yundi and Hager, Paul and Liu, Che and Shit, Suprosanna and Chen, Chen and Rueckert, Daniel and Pan, Jiazhen},
  journal={Medical Image Analysis},
  pages={103756},
  year={2025},
  publisher={Elsevier}
}

@article{ruijsink2020fully,
  title={Fully automated, quality-controlled cardiac analysis from CMR: validation and large-scale application to characterize cardiac function},
  author={Ruijsink, Bram and Puyol-Ant{\'o}n, Esther and Oksuz, Ilkay and Sinclair, Matthew and Bai, Wenjia and Schnabel, Julia A and Razavi, Reza and King, Andrew P},
  journal={Cardiovascular Imaging},
  volume={13},
  number={3},
  pages={684--695},
  year={2020},
  publisher={American College of Cardiology Foundation Washington DC}
}

@article{kong2022learning,
  title={Learning whole heart mesh generation from patient images for computational simulations},
  author={Kong, Fanwei and Shadden, Shawn C},
  journal={IEEE Transactions on Medical Imaging},
  volume={42},
  number={2},
  pages={533--545},
  year={2022},
  publisher={IEEE}
}

@article{yang2020unsupervised,
  title={Unsupervised MR-to-CT synthesis using structure-constrained CycleGAN},
  author={Yang, Heran and Sun, Jian and Carass, Aaron and Zhao, Can and Lee, Junghoon and Prince, Jerry L and Xu, Zongben},
  journal={IEEE transactions on medical imaging},
  volume={39},
  number={12},
  pages={4249--4261},
  year={2020},
  publisher={IEEE}
}

@article{bargshady2022application,
  title={Application of CycleGAN and transfer learning techniques for automated detection of COVID-19 using X-ray images},
  author={Bargshady, Ghazal and Zhou, Xujuan and Barua, Prabal Datta and Gururajan, Raj and Li, Yuefeng and Acharya, U Rajendra},
  journal={Pattern Recognition Letters},
  volume={153},
  pages={67--74},
  year={2022},
  publisher={Elsevier}
}

@article{ozbey2023unsupervised,
  title={Unsupervised medical image translation with adversarial diffusion models},
  author={{\"O}zbey, Muzaffer and Dalmaz, Onat and Dar, Salman UH and Bedel, Hasan A and {\"O}zturk, {\c{S}}aban and G{\"u}ng{\"o}r, Alper and Cukur, Tolga},
  journal={IEEE Transactions on Medical Imaging},
  volume={42},
  number={12},
  pages={3524--3539},
  year={2023},
  publisher={IEEE}
}

@inproceedings{li2023bbdm,
  title={Bbdm: Image-to-image translation with brownian bridge diffusion models},
  author={Li, Bo and Xue, Kaitao and Liu, Bin and Lai, Yu-Kun},
  booktitle={IEEE/CVF conference on computer vision and pattern recognition},
  pages={1952--1961},
  year={2023}
}

@article{arslan2025self,
  title={Self-consistent recursive diffusion bridge for medical image translation},
  author={Arslan, Fuat and Kabas, Bilal and Dalmaz, Onat and Ozbey, Muzaffer and {\c{C}}ukur, Tolga},
  journal={Medical Image Analysis},
  volume={106},
  pages={103747},
  year={2025},
  publisher={Elsevier}
}

@inproceedings{chadebec2025lbm,
  title={Lbm: Latent bridge matching for fast image-to-image translation},
  author={Chadebec, Cl{\'e}ment and Tasar, Onur and Sreetharan, Sanjeev and Aubin, Benjamin},
  booktitle={Proceedings of the IEEE/CVF International Conference on Computer Vision},
  pages={29086--29098},
  year={2025}
}
\end{document}